\documentclass[conference]{IEEEtran}
\IEEEoverridecommandlockouts
\usepackage{cite}
\usepackage{amsmath,amssymb,amsfonts}
\usepackage{graphicx}
\usepackage{textcomp}
\usepackage{xcolor}
\usepackage{lineno}					
\usepackage{tikz} 					
\usepackage{float}
\usepackage{array}
\usepackage{algorithm}
\usepackage{algpseudocode}
\usepackage{multirow}
\usepackage{subcaption}
\def\BibTeX{{\rm B\kern-.05em{\sc i\kern-.025em b}\kern-.08em
    T\kern-.1667em\lower.7ex\hbox{E}\kern-.125emX}}
\begin{document}

\title{Interpretable Sequence Clustering\\
}

\author{
\IEEEauthorblockN{Junjie Dong}
\IEEEauthorblockA{\textit{School of Software} \\
\textit{Dalian University of Technology}\\
Dalian, China \\
jd445@qq.com}
\and
\IEEEauthorblockN{Xinyi Yang}
\IEEEauthorblockA{\textit{School of Software} \\
\textit{Dalian University of Technology}\\
Dalian, China \\
yxinyi1209@163.com}
\and
\IEEEauthorblockN{Mudi Jiang}
\IEEEauthorblockA{\textit{School of Software} \\
\textit{Dalian University of Technology}\\
Dalian, China \\
792145962@qq.com}
\and
\IEEEauthorblockN{Lianyu Hu}
\IEEEauthorblockA{\textit{School of Software} \\
\textit{Dalian University of Technology}\\
Dalian, China \\
hly4ml@gmail.com}
\and
\IEEEauthorblockN{Zengyou He}
\IEEEauthorblockA{\textit{School of Software} \\
\textit{Dalian University of Technology}\\
Dalian, China \\
zyhe@dlut.edu.cn}
}

\maketitle

\begin{abstract}

Categorical sequence clustering plays a crucial role in various fields, but the lack of interpretability in cluster assignments poses significant challenges. Sequences inherently lack explicit features, and existing sequence clustering algorithms heavily rely on complex representations, making it difficult to explain their results.
To address this issue, we propose a method called Interpretable Sequence Clustering Tree (ISCT), which combines sequential patterns with a concise and interpretable tree structure. ISCT leverages $k-1$ patterns to generate $k$ leaf nodes, corresponding to $k$ clusters, which provides an intuitive explanation on how each cluster is formed.
More precisely, ISCT first projects sequences into random subspaces and then utilizes the $k$-means algorithm to obtain high-quality initial cluster assignments. Subsequently, it constructs a pattern-based decision tree using a boosting-based construction strategy in which sequences are re-projected and re-clustered at each node before mining the top-$1$ discriminative splitting pattern.
Experimental results on 14 real-world data sets demonstrate that our proposed method provides an interpretable tree structure while delivering fast and accurate cluster assignments.

\end{abstract}


\section{Introduction}
A categorical sequence is a series of symbols arranged in a specific order, where the symbols belong to a finite set of items. Categorical sequence clustering is an unsupervised machine-learning problem that aims to partition unlabelled sequences into homogeneous groups. This technique is widely used in several fields, including bioinformatics \cite{zou2020sequence}, natural language processing \cite{4408578}, and financial data analysis \cite{gupta2020comprehensive}. For instance, in bioinformatics, sequence clustering can aid in identifying gene families and determining the evolutionary relationships between different species. 

Existing research efforts mainly focus on how to obtain accurate clustering results rather than the interpretation of identified clusters. For scientific purposes, it is crucial to understand the reasons why certain genes are grouped into particular families and what drives the model to make these divisions. However, the current sequence clustering methods \cite{ranjan_sequence_2021,bose2009context,guralnik2001scalable,society2013novel,dinu2014clustering,chen2017sequence,karatzoglou2010kernel} simply output the clusters of sequences and none of them provides a concise explanation of the cluster assignments, as elaborated below. 


A significant amount of algorithms for clustering sequences have been proposed during the last decades. These methods can be roughly divided into feature/pattern-based methods, partitional methods, hierarchical methods, and  model-based methods. However, the cluster assignments in these methods are typically hard to explain in a concise manner that is comprehensible to humans. 

Firstly, feature/pattern-based methods \cite{guralnik2001scalable,ranjan_sequence_2021} transform the sequence into a feature vector based on extracted patterns and then use vectorial data clustering methods like $k$-means for clustering. Although the first step retains some interpretability by converting the sequence into numerical vectors, the subsequent vectorial data clustering step is highly complex. For instance, in $k$-means, each input vector is assigned to a cluster based on its distance to the center of each cluster. However, since each feature contributes to the distance calculation, it is challenging to discern a clear rationale for the decision process.

Partitional methods \cite{chen2017sequence} divide sequences into several clusters by optimizing an objective function, while hierarchical methods \cite{bose2009context} use distance measures to obtain a hierarchical structure. However, these methods do not produce a human-friendly structure that directly reveals the decision process. Model-based methods \cite{society2013novel} assume that sequences from different clusters have different distributions, and then assign sequences to clusters accordingly. However, for those unfamiliar with statistical techniques, this approach makes it even more difficult to understand why certain sequences are allocated to the same cluster.

In recent years, there has been an increasing research interest in interpretable clustering methods for non-sequential data \cite{basak_interpretable_2005,bandyapadhyay_how_2022,bertsimas_interpretable_2021,carrizosa_interpreting_2022,moshkovitz2020explainable,fraiman_interpretable_2011,lawless_interpretable_2022,makarychev_explainable_2022}. These methods can generally be classified into two main categories: \textit{pre}-model and \textit{post}-model methods. The focus of the \textit{post}-model approach is to give reasoning behind the decision made by the black box model, i.e., to explain why the model partitions data into certain clusters. In contrast, the \textit{pre}-model approach emphasizes the development of a concise model that is easily understood or perceivable by humans, with an intuitively simple structure. The most representative model of this type is the threshold tree \cite{moshkovitz2020explainable}.
 

The threshold tree is an interpretable clustering method that divides the data set into several partitions with a binary tree structure. Each node in the tree iteratively partitions the data based on a threshold value derived from a specific feature. This hierarchical clustering approach allows for interpretable clustering results. Unfortunately, the current state-of-the-art interpretable clustering algorithms are mainly focused on continuous vector data and cannot be directly applied to sequential data. This is mainly because the features of the sequential data are hidden and the potential dimensionality is very high.

Inspired by the threshold tree, we propose a new explainable clustering structure for sequential data based on patterns, which offers a concise way for humans to understand the partition. Interpretable Sequence Clustering Tree (ISCT) employs a binary tree structure and is built in a top-down manner. The splitting criterion is based on the presence of certain discriminative sequential patterns, which are selected based on both their discriminative power and their support.  The main steps for constructing an unsupervised pattern tree are as follows. First, we use sequence clustering algorithms to obtain initial cluster assignments. Then, we perform discriminative sequential pattern mining on the sequential data based on the partition information to obtain the most discriminative sequential pattern. We consider sequences that contain the same pattern to belong to the same cluster, and we repeat the process for the remaining $k-1$ clusters. Through this iterative process, we can obtain a hierarchical unsupervised pattern tree with $k-1$ nodes.

The contribution of this paper is as follows:
\begin{itemize}
    \item   A tree-based interpretable sequence clustering method named Interpretable Sequence Clustering Tree is proposed in this paper.  To our knowledge, this is the first interpretable clustering method for sequential data.
    \item    ISCT algorithm constructs a tree in a boosting manner by utilizing a small number of discriminative patterns, resulting in a clear and interpretable structure. This provides a concise and understandable way for humans to interpret the clustering results.
    \item The experimental results on 14 real-world data sets show that the proposed ISCT can achieve comparable performance to state-of-the-art methods while providing a clear and interpretable structure.
\end{itemize}





 

\section{Related Work}
\label{relatedwork}
This section discusses previous works closely related to our approach. A brief overview of the current advancements in the field of interpretable clustering is presented in Section \ref{interpretable Clustering}, while the latest research on sequence clustering is described in Section \ref{sequence clustering}. Additionally, Section \ref{disseqming} provides an introduction to some typical methods for discriminative sequential pattern mining.

\subsection{Interpretable Clustering}
\label{interpretable Clustering}

\subsubsection{Fuzzy Rule-based Approach}
Fuzzy rule-based methods aim to extract fuzzy IF-THEN rules from data, which take the form:

\centerline{ Rule $R_i$: IF $x_1$ is $\phi_{11}$ and $\ldots$ and $x_n$ is $\phi_{jj}$ }
\centerline{THEN cluster $c_i$.}

Here, $x_n$ represents the $n$-th feature of a sample $x$, and $\phi_{jj}$ is the fuzzy number that must be satisfied for that feature. By discovering different rules that are satisfied by one or more samples within a cluster, the decision process of the clustering algorithm can be explicitly characterized with high interpretability. Fuzzy rule-based methods can be broadly classified into the adaptive partition and fixed grid partition based on whether the membership function is related to the data \cite{yang2021survey}.

The information of data is taken into account during the process of the adaptive partition method. For example, FRCGC \cite{wang2014rapid} is a rapid fuzzy rule clustering method based on granular computing. It provides descriptions for all clusters by using exemplar descriptions selected from sample descriptions to guide data granulation, resulting in each cluster being depicted by a single fuzzy rule. For fixed grid partition, the information of the data set is not considered in the fuzzy partition process. Commonly used membership functions in this context are the trigonometric and trapezoidal ones \cite{mansoori2011frbc,hsieh2016gminer}.

\subsubsection{Polytope-based Approach}
The polytope-based approach uses hyperplanes as the boundaries of the clusters, which are combined to form a polytope. The description of the hyperplanes is then used to construct rules that enable interpretable clustering. This approach is similar to Support Vector Machines (SVM) since they both use hyperplanes as the decision boundary. However, unlike SVM, the polytope-based clustering method does not rely on labelled data, making the task of finding the hyperplanes for the clustering boundary more challenging. 

From this perspective, many different algorithms have been proposed. For example, in \cite{pelleg2001mixtures}, hyperrectangle boxes are used as the clustering boundary. Initially, the hyperrectangles may overlap, and to obtain the final rectangles, a soft ``tail" is introduced to calculate the membership of samples to different rectangles. The algorithm then calculates the final upper and lower boundaries using an alternating minimization strategy. The Discriminative Rectangle Mixture Model (DReaM) \cite{chen2016interpretable}, on the other hand, is a more flexible approach as it allows for the specification of two types of features: rule-generating features and cluster-preserving features. Specifically, DReaM constructs rules for clustering using only the rule-generating features, while finding the clustering structure using the cluster-preserving features. Unlike the rectangle decision boundaries used in the previous two methods, \cite{lawless_interpretable_2022} constructs polytopes around each cluster and translates the decision boundaries into rules to obtain interpretable clustering results.

\subsubsection{Threshold Tree-based Approach }
Tree-based methods are widely used in the fields of classification and clustering due to their simplicity and intuitive interpretability. In the field of clustering, threshold tree methods can be divided into two classes based on whether they require the label information of other clustering algorithms.


Four different attribute selection metrics and two data partitioning algorithms are proposed in \cite{basak_interpretable_2005}, which introduces a hierarchical clustering method for generating interpretable tree structures. In contrast to the direct construction of the tree as done previously, CUBT \cite{fraiman_interpretable_2011, ghattas2017clustering} incorporates random auxiliary samples as a second class in the sample space to acquire the necessary labels for calculating information gain. To enhance the algorithm's performance, the decision process is optimized by adjusting the splitting criterion.

%
The tree construction process can also leverage pseudo-labels generated by conventional clustering methods \cite{moshkovitz2020explainable,gamlath2021nearly,bandyapadhyay_how_2022,makarychev_explainable_2022}. These studies tackled this problem from a theoretical standpoint and evaluated cluster quality using the $k$-means and $k$-medians objectives. Especially, the performance of entropy-based decision tree algorithms with threshold cut algorithms are compared in \cite{moshkovitz2020explainable}, demonstrating that threshold cut achieves a constant factor approximation for both means and medians. Subsequent research has focused on improving the construction strategy to minimize the cost between the constructed threshold tree and the $k$-means algorithm. 
There are similarities between the work presented in this paper and the aforementioned studies, as both adopt the approach of constructing a smaller-scale decision tree to achieve a highly interpretable decision outcome.


\subsection{Categorical Sequence Clustering}
\label{sequence clustering}

\subsubsection{Feature/Pattern-Based Method}
\label{featurebased}
There are primarily two different strategies for the feature or pattern-based methods. The first approach involves transforming the sequence into a feature vector and then applying existing clustering methods such as $k$-means \cite{yuan2019two,guralnik2001scalable} or DBSCAN \cite{li2020ssrdvis} to obtain the final partition results. The differences between these methods mainly arise from the process of converting sequences into feature vectors. Popular approaches include using sequential patterns \cite{guralnik2001scalable,li2012efficient}, $k$-mers \cite{steinegger2018clustering}, or embedding techniques \cite{ranjan_sequence_2021,li2020ssrdvis}.
Another strategy involves initially constructing a clustering result based on the discovered sequential patterns and subsequently expanding and optimizing these clusters to achieve the final refined clustering results.

\subsubsection{Model-based Method}
Model-based methods are based on the assumption that sequences originating from distinct clusters follow separate probability distributions. Existing model-based approaches utilize either Markov models, constructing Markov chains that consider the occurrence of a sequence category given its preceding subsequences \cite{society2013novel,xiong2011new,yang2003cluseq,cadez2003model}, or the Hidden Markov Model (HMM), which introduces hidden states to the sequences \cite{smyth1996clustering,bicego2003similarity}.

\subsubsection{Hierarchical Method}

Hierarchical methods primarily rely on the distances between all sequences. Typically, pairwise similarity matrices need to be computed as a first step. Subsequently, either agglomerative or divisive approaches are employed to progressively build the hierarchical structure. The main distinction between these methods lies in their utilization of different distance functions to obtain the similarity matrix. Commonly used distance functions include Kullback-Leibler distance \cite{ramoni2002bayesian}, edit distance \cite{bose2009context}, and Jaccard similarity based on subsequences \cite{oh2004hierarchical}. 

It is worth mentioning that this method already exhibits a certain level of interpretability, as the partition can be explained based on the distances between samples. However, this form of explanation may not be intuitive enough. The computation of distances is not straightforward for humans, as it involves complex combinations of individual elements within the sequences.

\subsubsection{Partitional Method}
Partitional methods \cite{chen2017sequence} directly divide a set of sequences into multiple clusters. Instead of transforming the sequence vectors and applying traditional partitional clustering algorithms introduced in Section \ref{featurebased}, these methods aim to obtain partition results directly in the original sequence space.

For example, a modified $k$-Median algorithm is presented in \cite{dinu2014clustering}. It adopts the rank-based distance to replace the edit distance for accelerating the computational time. Thus, each cluster can be presented by a centroid sequence.

\subsection{Discriminative Sequential Pattern Mining}
\label{disseqming}
Discriminative sequential pattern mining is closely related to sequence classification \cite{xing_brief_2010}. Its objective is to discover patterns that exhibit strong correlations with specific labels. These mined patterns are characterized by their frequent occurrence in sequences with one label and their absence as subsequences in sequences with another label. The identification of discriminative patterns for sequence classification has been extensively investigated, with each study focusing on different measurement strategies  \cite{ji_mining_2007,fradkin2015mining,zhou_pattern_2016,de_smedt_mining_2020}.
The cSPADE method was enhanced by incorporating an interestingness measure that considers both the support and the window (cohesion) in which the constraint items occur \cite{zhou_pattern_2016}. Furthermore, BIDE was extended to BIDE-D(C) by incorporating information gain, providing a direct pattern mining approach to sequence classification \cite{fradkin2015mining}. On the other hand, iBCM specifies a series of templates and filters them based on whether the sequences in the hit only exist in sequences of a single category. This approach aims to obtain a compact set of informative patterns \cite{de_smedt_mining_2020}.

\section{Problem Statements}
\label{satement}
\subsection{Notation}
Given a set of items $ \mathcal{I} = \left\{e_1 , e_2 , \dots , e_m\right\}$ with $m$ distinct items, a discrete sequence $s =\langle \sigma_1 , \sigma_2 , \dots , \sigma_l \rangle$ with length $|s| = l$ is an ordered list over $\mathcal{I}$ if each $\sigma_i \in \mathcal{I}$.  A subsequence $t$ of $s$ is defined as a sequence that can be obtained by removing some elements from $s$ while preserving the relative order of the remaining elements. Formally, $t$ is a subsequence of $s$, denoted as $t \subseteq s$, if there exists a strictly increasing sequence of indices $1 \leq i_1 < i_2 < \dots < i_k \leq l$ such that $t = \langle \sigma_{i_1}, \sigma_{i_2}, \dots, \sigma_{i_k} \rangle$.

\subsection{Support}
A sequence database $\mathcal{D}$ is a list of sequences, where the number of sequences in the database is denoted as $|\mathcal{D}|$. The number of occurrences of a given sequential pattern $p$ in the database is denoted as $Occ(p,\mathcal{D}) $. The support of a pattern $p$ in $\mathcal{D}$ is defined as:
\begin{equation}
    Supp(p,\mathcal{D}) = \frac{Occ(p,\mathcal{D}) }{|\mathcal{D}|}.
    \label{eq:support}
\end{equation}

Given a threshold $\alpha_{Supp}$, a sequential pattern $p$ can be regarded as a frequent sequential pattern if  $Supp(p,\mathcal{D}) \geq \alpha_{Supp} $.

However, directly specifying the support threshold can be challenging in many cases. As a practical alternative, it is common to set a parameter $k$ to obtain the top-$k$ patterns with sufficiently high support.


\subsection{Problem Formulation}
 For a given database $\mathcal{D}$  and an integer $k$, the sequence clustering method $\mathcal{T}$ partitions the database $\mathcal{D}$ into $k$ subsets $C = \{c_1 , c_2 , \dots , c_k\}$. Our goal is to construct an interpretable sequence clustering tree characterized by its compact and concise structure, enabling intuitive and comprehensive understanding by human users. In detail, the clustering tree should exhibit the following characteristics:
 \begin{itemize}
     \item \textit{Shallow}: the interpretability of the decision tree is determined by its size. Similar to \cite{moshkovitz2020explainable}, our objective is to obtain a tree with exactly $k$ leaves, one for each cluster $\{1,2, \dots,k \}$, which means that the depth and the total number of features to at most $k-1$.
     \item \textit{Concise}: the splitting criteria for trees should be concise, particularly in the case of sequential data where we aim to avoid incorporating distance functions or similarity matrices in the splitting process. Similar to vector data, the calculation of distances involves the interrelationship of multiple items.
\item \textit{Consistency}: The interpretable clustering tree should accurately represent and approximate the original result. This involves two aspects: firstly, ensuring the number of partitions in the clustering tree is consistent with the original results; secondly, achieving competitive performance compared to the original clustering method.

 \end{itemize}

\section{Interpretable Sequence Clustering Tree}
\label{themethod}
\subsection{An Overview of the Algorithm}
In order to construct an Interpretable Sequence Clustering Tree (ISCT) for a given sequence database, we can utilize cluster membership as a pseudo label and apply various tree-based classification methods. However, it is important to note that the interpretability of a decision tree largely depends on its size. Unfortunately, using these tree-based methods directly often results in complex and non-interpretable tree structures, as these methods tend to generate large and complex trees. Even with control over splitting procedures or the use of pruning strategies, creating an intuitive and concise decision tree remains a challenging task in practice.

\begin{figure*}[!htbp]%
    \centering
    \includegraphics[scale = 0.45]{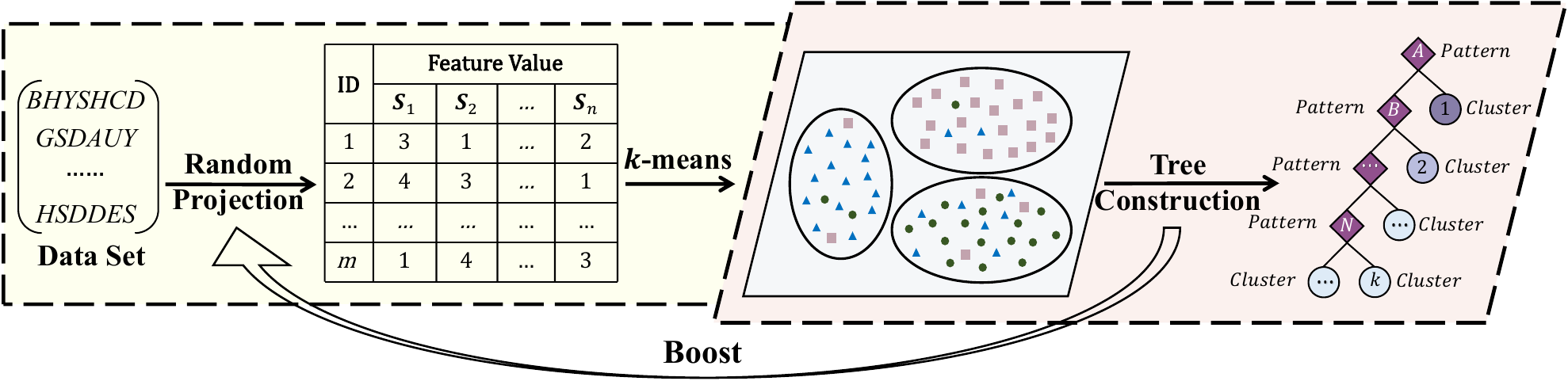}
    \caption{The workflow of the Interpretable Sequence Clustering Tree. It mainly consists of two parts: (1) Clustering via random projection and (2) Tree construction.}
    \label{The workflow of Interpretable Sequence Clustering Tree}
\end{figure*}

The most intuitive approach to separate two clusters in a sequence database is by utilizing the presence or absence of a specific sequential pattern \cite{9411677}. Our objective is to construct a decision tree with $k$ leaf nodes and $k-1$ patterns, where each leaf node comprises sequences belonging to the same cluster. The workflow of our method is illustrated in Figure \ref{The workflow of Interpretable Sequence Clustering Tree}. Our method consists of two main components. The first component of our method involves rapid vectorization by randomly projecting sequences into subspaces, facilitating the efficient utilization of the $k$-means algorithm for obtaining initial partition results. This component is elaborated on in Section \ref{Random Subsequence Projection Clustering}. The second component centers around the tree construction algorithm, as explained in Section \ref{construction}. In this section, we introduce a novel measure based on relative risk to identify discriminative patterns and outline the procedure for constructing the tree based on the top-1 discriminative pattern obtained.

\begin{algorithm}[htbp]
    \caption{Random Projection Clustering}
    \label{Random Projection Clustering}
    \begin{algorithmic}[0]
        \State \textbf{Input:} Sequence database $\mathcal{D} = \{s_1,s_2,\dots,s_n \}$, number of clusters $k$, pattern number $numP$, maximal random pattern length $maxS$.
        \State \textbf{Output:} Clusters $C = \{c_1, c_2, \dots c_k \}$
    \end{algorithmic}
    \begin{algorithmic}[1]

        \State $\mathcal{I} \gets readItem(\mathcal{D})$
        \State $Pset \gets \emptyset$

        \For{$i \in [1, numP]$}
        \State $l = random(1, maxS)$
        \State $p = randomSubsequence(l, \mathcal{I})$
        \State $Pset.add(p)$
        \EndFor
        \State $feature \gets lcsTransform(\mathcal{D}, Pset)$
        \State $feature \gets PCA(feature, k)$
        \State $C \gets kmeans(feature, k)$
        \State \textbf{return} $C$
    \end{algorithmic}

\end{algorithm}
\subsection{Clustering via Random Projections}
\label{Random Subsequence Projection Clustering}
Unlike most feature-based methods that rely on frequent pattern mining or sequence embeddings, our approach takes a different perspective. We propose a method that utilizes randomly generated sequential patterns to transform sequences into vectors. That is, by calculating the distance between each random pattern and the target sequence, each sequence in the database is projected into the corresponding subspace.

In this paper, we employ the longest common subsequence (LCS) to compute the distance between two sequences (the random sequential pattern and each sequence in the database). To mitigate the impact of redundant patterns and enhance stability, we further employ PCA (Principal Component Analysis) for dimensionality reduction. This technique reduces the dimensionality of the feature vectors to a lower-dimensional space. Subsequently, we apply the $k$-means algorithm to cluster the transformed vectorial data.

The overall procedure is outlined in Algorithm \ref{Random Projection Clustering}. The main steps are as follows: First, we randomly generate a number of subsequences from the sequence database $\mathcal{D}$ (lines 3 to 6). Then, we transform the sequences into feature vectors using the LCS similarity (line 8). Subsequently, PCA is used for conducting the dimension reduction (line 9). Finally, we cluster the feature vectors using $k$-means to obtain the clustering result (line 10).

\subsection{Tree Construction}
\label{construction}

We can utilize the existence of a specific pattern as the splitting criterion for building the tree. Specifically, given initial clusters $C = \{ c_1, c_2, \dots, c_k \}$, the set of clusters delivered by the decision tree is $\hat{C} = \{\hat{c}_1,\hat{c}_2, \dots, \hat{c_k} \}$. In order to construct the tree, we need to find representative patterns for each of the $k-1$ clusters. During the construction of the clustering tree, the database $\mathcal{D}$ is firstly divided into two sets $\mathcal{D}_{p}$ and $\overline{\mathcal{D}_{p}}$ based on the existence of a sequential pattern $p$. The problem is then converted into finding the top-1 discriminative pattern $p$ for tree splitting.

However, directly using traditional splitting criteria such as the Gini impurity may not perform well when dealing data sets with a large number of clusters. In this paper, we present a new split measure based on relative risk for identifying discriminative patterns.

\begin{algorithm}[htbp]
    \caption{ISCT Construction}
    \label{ISCTconstruction}
    \begin{algorithmic}[0]
        \State \textbf{Input:} Sequence database $\mathcal{D} = \{s_1,s_2,\dots,s_n \}$, number of clusters $k$, indicator of boosting $boost$,
         maximal pattern number $maxN$, minimum number of sequences $minS$.
        \State \textbf{Output:} root of the pattern tree $N$
    \end{algorithmic}
    \begin{algorithmic}[1]
        \State $C = \{c_1 , c_2 , \dots , c_k\} \gets Algorithm \  \ref{Random Projection Clustering}(\mathcal{D},k)$ 
        \State $N \gets creatNode(\mathcal{D})$
        \State $Built\_Tree(N,\mathcal{D},C)$
        \State \textbf{return} $N$

    \end{algorithmic}
    \begin{algorithmic}[1]
    \Function{Built\_Tree}{$N,\mathcal{D},C$}
        \If{$k < 2$ \textbf{or} $|\mathcal{D}| \leq min(K,minS)$}
            \State \textbf{return}
        \EndIf
        \If{$boost = True$}
            \State $C \gets Algorithm \  \ref{Random Projection Clustering}(\mathcal{D},k)$
        \EndIf
        \State $candidatePatterns \gets \emptyset$
        \For{each cluster $c_i \in C$}
        \State $freqPattern \gets mineFreqPattern(c_i,maxN)$
            \State $candidatePatterns.add(freqPattern)$
        \EndFor
        \State $p \gets getTop1Pattern(candidatePatterns)$
        \If{$p = \emptyset$}
            \State \textbf{return}
        \EndIf
        \State $\mathcal{D}_{p} \gets \{s \in \mathcal{D} | p \subseteq s \}$
        \State $\overline{\mathcal{D}_{p}} \gets \{s \in \mathcal{D} | p \not\subseteq s \}$
        \State $k \gets k - 1$
        \State $N.left \gets createNode(\overline{\mathcal{D}_{p}})$
        \State $N.right \gets createNode(\mathcal{D}_{p})$
        \State $Built\_Tree(N.left,\overline{\mathcal{D}_{p}},C)$
        \State $Built\_Tree(N.right,\mathcal{D}_{p},C)$

    \EndFunction
    \end{algorithmic}
\end{algorithm}

\textit{\textbf{Splitting criterion}} The relative risk (RR) is a commonly used measure for evaluating the discriminative power of sequential patterns. The relative risk of pattern $p$ is calculated as the ratio of the support in positive class $\mathcal{D}^+$ to the support in negative class $\mathcal{D}^-$:
\begin{equation}
RR = \frac{Supp(p,\mathcal{D}^+)}{Supp(p,\mathcal{D}^-)}.
\end{equation}
In addition to relative risk, the following internal similarity measure is employed to evaluate the goodness of a pattern $p$: 
\begin{equation}
SIM = \frac{|p| \times |\mathcal{D}^+|}{\sum_{j=1}^{|\mathcal{D}^+|} |s_i|},
\end{equation}
where $|p|$ is the length of pattern $p$, and $|\mathcal{D}^+|$ is the number of sequences in the positive class $\mathcal{D}^+$. That is, we prefer longer patterns if they have the same discriminative capability.

However, the concept of relative risk can only be applied as a measure when there are two classes involved. For output $k$ partitions, we need to extend the concept of relative risk to accommodate multiple classes. To address this requirement, $\mathcal{D}^+$ is defined as:

\begin{equation}
\mathcal{D}^{+}=\{c_i \mid \arg\max_{i} {Supp\left(p, c_i\right)}\}.
\end{equation}

We select the cluster with the highest support of pattern $p$ as the positive class $\mathcal{D}^{+}$ and obtain the negative class $\mathcal{D}^{-}$ using the one-vs-rest strategy, where $\mathcal{D}^{-} = \mathcal{D} \setminus \mathcal{D}^{+}$. We try to identify a pattern that exhibits the highest discriminative power according to Equation (2). If multiple patterns have the same relative risk, we choose the pattern among them with the maximal SIM value according to Equation (3). The presence of this pattern serves as the basis for partitioning the data set.

\textit{\textbf{Tree construction}} The top-$1$ discriminative sequential pattern is used as the splitting point for constructing the tree, where the sequences containing this pattern are assigned to the same cluster. Based on this idea, we propose an algorithm for building an interpretable sequence clustering tree.

The construction process is described in Algorithm \ref{ISCTconstruction}. The first step is to obtain $k$ initial clusters based on Algorithm \ref{Random Projection Clustering}. After initializing the root node $N$, we call the $Built\_Tree$ to construct the pattern tree.

The tree construction follows a top-down approach with binary splits. In the first stage, the algorithm checks the stop criteria of the given node (line 2 to 4). The first criterion is $k < 2$, which determines whether the desired number of leaf nodes has been reached. The second criterion is $|\mathcal{D}| \leq \min(k, minS)$, which determines whether to continue splitting based on the minimum number of sequences included.

After checking the stop criteria, a frequent sequential pattern mining algorithm is applied to extract frequent sequential patterns for each $c_i \in C$ (line 9 to 12). The top-$1$ discriminative pattern $p$ is selected using different measures (line 13). If a sequence $s$ in $\mathcal{D}$ contains pattern $p$, it is assigned to the right node. On the other hand, if a sequence does not contain pattern $p$, it is assigned to the left node and the algorithm continues building the tree by recursively calling the $Built\_Tree$ function (line 17 to 23). The function is terminated when all the nodes in the tree satisfy the stop criteria.


\textit{\textbf{Boost Construction}} In our experiments, we have observed that this direct construction method often leads to inconsistencies between the reported clusters $\hat{C}$ and given partition $C$ when dealing with data sets with more clusters. This inconsistency arises from the difference between the cluster $\hat{c_i}$ constructed using the splitting pattern and the expected cluster $c_i$ given by Algorithm 1. 

\begin{figure}[!t]%
    \centering
    \includegraphics[scale = 0.55]{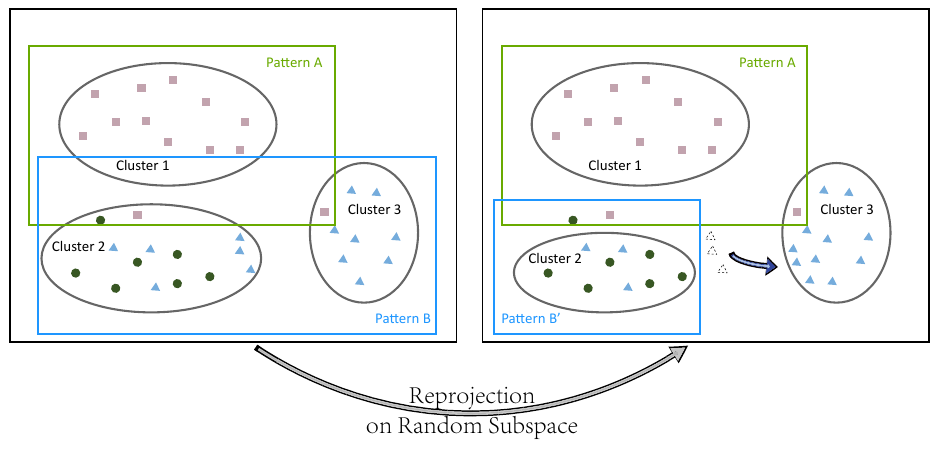}
    \caption{The inconsistency between the reported clusters and given partitions and the boosting strategy.}
    \label{boostwhy}
\end{figure}

To address this problem, we propose a boosting strategy during the tree construction process. In other words, each $C$ is re-generated using Algorithm 1 based on the current remaining cluster number $k$. Figure \ref{boostwhy} illustrates this problem and the boosting result.

Due to the limitations of random projection clustering, the resulting sequence clusters may not always be accurate. For instance, in Figure \ref{boostwhy}, cluster 2 is not pure as it contains sequences from both cluster 1 and cluster 3. However, when we remove $\mathcal{D}_p$ from the database, some items that only appear in sequences of cluster 1 are also discarded. To utilize this information, we propose a boosting strategy to re-partition the database $\mathcal{D}$ by randomly re-projecting the sequences into different subspaces. This approach can partially calibrate our results and improve their consistency with the base sequence clustering method.

\begin{figure}[!t]%
    \centering
    \includegraphics[scale = 0.45]{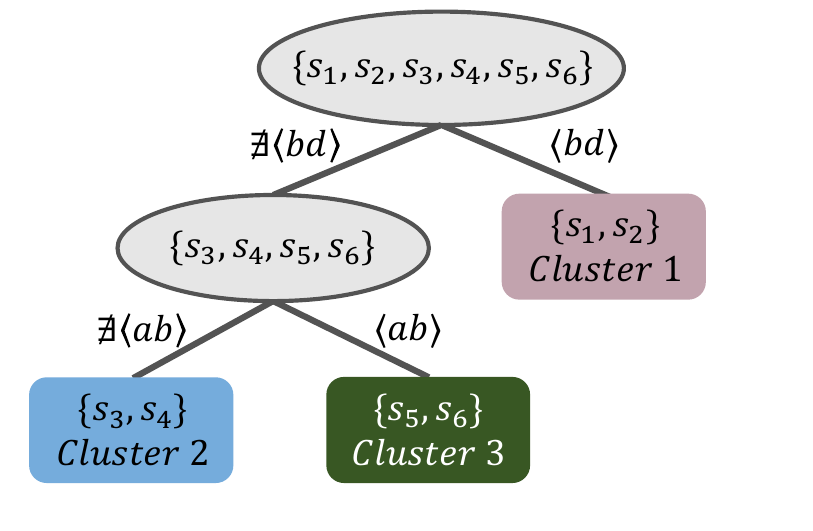}
    \caption{The structure of the pattern tree: Each splitting node represents a sequential pattern.}
    \label{PatternTree}
\end{figure}

\textit{\textbf{Example 1.}} Consider the toy example sequence database presented in Table \ref{toy_database}, which consists of six sequences ($|\mathcal{D}| = 6$) and a set of five items denoted as $\mathcal{I} = \{a,b,c,d,e\}$. The database is partitioned into three clusters. Figure \ref{PatternTree} illustrates the construction of a pattern tree for Example 1.

\begin{table}[!t]
\centering
\caption{Toy sequence database and the clustering result}
\begin{tabular}{|c||c|c||c||c|c|}
\hline
ID & Sequence & Cluster & ID & Sequence  & Cluster \\ \hline\hline
$s_1$  & $\langle abddc\rangle$   & 1     & $s_2$  & $\langle adbbded\rangle$   & 1     \\\hline
$s_3$  & $\langle acdeeaa\rangle$  & 2     & $s_4$  & $\langle acaeadcd\rangle$   & 2     \\\hline
$s_5$  & $\langle abbcaac\rangle$   & 3     & $s_6$  & $\langle acbbccaa\rangle$   & 3     \\ \hline
\end{tabular}
\label{toy_database}
\end{table}

The first selected top-$1$ discriminative pattern is $p_1 = \langle bd \rangle$, which splits the data into two clusters. The first cluster $\hat{c_1}$ is obtained, containing sequences $s_1$ and $s_2$. The remaining sequences are then assigned to the left node. Since the stop criteria indicate that the left node should be further divided, the top-$1$ discriminative pattern $p_2 = \langle ab \rangle$ is selected. Eventually, we obtain a highly concise and interpretable clustering tree with three leaf nodes, represented by rectangles in the figure.




\section{Experiments}
\label{experimentsonisct}
We conduct extensive experiments on 14 real-world data sets to evaluate the performance of ISCT. Specifically, we aim to answer the following Research Questions (RQs):

\begin{itemize}
\item \textbf{RQ1:} How does our method compare to state-of-the-art approaches in terms of identification performance?

\item \textbf{RQ2:} What is the performance difference between using a naive tree-based approach and ISCT in terms of interpretability?
\item \textbf{RQ3:} What are the advantages of using the Boost method in constructing interpretable clustering trees?
\end{itemize}

\subsection{Experimental Setup}
    
\subsubsection{Data Sets}

We used 14 real-world data sets to evaluate the performance of our algorithm: Activity \cite{asuncion2007uci}, Aslbu \cite{di2013two}, Auslan2 \cite{di2013two}, Context \cite{maggi2012efficient}, Epitope \cite{maggi2016semantical}, Gene \cite{gama2014survey}, News \cite{zhang2015ccspan}, Pioneer \cite{di2013two}, Question \cite{lam2014mining}, Robot \cite{zhang2015ccspan}, Skating \cite{di2013two}, Reuters \cite{zhang2015ccspan}, Unix \cite{zhang2015ccspan}, and Webkb \cite{zhang2015ccspan}.
The details of these data sets are summarized in Table \ref{data set}, where $|\mathcal{D}|$ refers to the number of sequences, $|\mathcal{I}|$ is the number of items, min$l$ and max$l$ are the minimum and maximum length of the sequences, respectively.
\begin{table}[!t]
    \centering
    \caption{Some characteristics of the data sets used in the performance comparison.}
    \label{data set}
    \begin{tabular}{|c||c|c|c|c|c|}
        \hline Data sets & $|\mathcal{D}|$ & $|\mathcal{I}|$ & min$l$ & max$l$ & \#classes \\
        \hline\hline
        Activity       &35     &10     &12      &43      &2    \\\hline
        Aslbu          & 424   & 250   & 2      & 54     & 7   \\\hline
        Auslan2        & 200   & 16    & 2      & 18     & 10  \\\hline
        Context        & 240   & 94    & 22     & 246    & 5   \\\hline
        Epitope        & 2392  & 20    & 9      & 21     & 2   \\\hline
        Gene           & 2942  & 5     & 41     & 216    & 2   \\\hline
        News           & 4976  & 27884 &1       &6779    & 5   \\\hline
        Pioneer        & 160   & 178   & 4      & 100    & 3   \\\hline
        Question       & 1731  & 3612  & 4      & 29     & 2   \\\hline
        Robot          & 4302  & 95    & 24     & 24     & 2   \\\hline
        Skating        & 530   & 82    & 18     & 240    & 7   \\\hline
        Reuters        & 1010  & 6380  & 4      & 533    & 4   \\\hline
        Unix           & 5472  & 1697  & 1      & 1400   & 4   \\\hline
        Webkb          & 3667  & 7736  &1       &20628   & 3   \\\hline
    \end{tabular}
\end{table}

\subsubsection{Evaluation Measures}
The evaluation measures used in this study include Purity, NMI (Normalized Mutual Information), and F1-score.


\subsubsection{Baseline Methods}
The following baseline methods are included in the performance comparison:
\begin{itemize}
    \item Hierarchical clustering methods: HC \cite{bose2009context} adopts the standard edit distance to obtain the pairwise dissimilarity and average linkage is used to compute the distance between clusters. For MCSC \cite{society2013novel}, we utilize the B\_MCSC variant.
    \item Feature-based methods: For FB-LL \cite{guralnik2001scalable}, the minimum support is fixed at $0.05$, and the length of the pattern is constrained between $3$ and $5$. For SGT \cite{ranjan_sequence_2021}, we set $kappa=1$ and $lengthSensitive=False$.
    \item Partitional Method: kkmeans \cite{xu2022multi} uses the 3-spectrum string kernel. For $k$-Median \cite{dinu2014clustering}, the edit distance is used as the pairwise dissimilarity measure. For MinDL \cite{chen2017sequence}, the penalty parameters $\alpha$ and $\gamma$ are specified as $0.1$ and $0$, respectively.
\end{itemize}

\subsubsection{Implementation}
In the experiments, the number of final clusters $k$ is set to be the number of ground-truth clusters except for MinDL. The experiments were performed on a PC with an AMD Ryzen 9 5900X CPU with 32 GB memory, and the results were obtained by averaging ten trials. For our algorithm, we set the maximum pattern number to $2048$, and the default maximum pattern length to $5$ if min$l$ $\leq 10$, maximum pattern length is min$l$. The number of top-$k$ frequent patterns is set to $512$. Our source code is available online \footnote{https://anonymous.4open.science/r/Interpretable-Sequence-Clustering-Tree-97F7}.

\subsection{Investigation of RQ1}

\begin{table*}[]
\centering
 \caption{Performance comparison of ISCT, HC, FB-LL, MCSC, $k$-Median, MinDL, kkmeans, and SGT.}
\begin{tabular}{cccccccccc}
\hline
\multicolumn{1}{c|}{Datasets}                                            & Evaluation & \textbf{ISCT}                    & HC                              & FB-LL                           & MCSC                            & k-Median & MinDL                           & kkmeans                         & SGT                             \\ \hline
\multicolumn{1}{c|}{\multirow{3}{*}{Activity}}                          & Purity     & {0.994}                           & 0.600                             & 0.643                           & 0.686                           & 0.695    & 0.600                             & 0.720                            & {1}     \\
\multicolumn{1}{c|}{}                                                   & NMI        & {0.973}                           & 0.038                           & 0.129                           & 0.118                           & 0.231    & 0.078                           & 0.225                           & {1}     \\
\multicolumn{1}{c|}{}                                                   & F1-score & {0.989}                           & 0.652                           & 0.613                           & 0.537                           & 0.620     & 0.515                           & 0.612                           & {1}     \\ \hline
\multicolumn{1}{c|}{\multirow{3}{*}{Aslbu}}                             & Purity     & 0.491                           & 0.380                            & 0.501                           & 0.373                           & 0.441    & {0.507} & 0.469                           & 0.432                           \\
\multicolumn{1}{c|}{}                                                   & NMI        & 0.218                           & 0.028                           & {0.220}  & 0.046                           & 0.139    & 0.177                           & 0.157                           & 0.164                           \\
\multicolumn{1}{c|}{}                                                   & F1-score & 0.317                          & {0.366} & 0.286                           & 0.183                           & 0.268    & 0.126                           & 0.257                           & 0.306                           \\ \hline
\multicolumn{1}{c|}{\multirow{3}{*}{Auslan2}}                           & Purity     & 0.348 & 0.195                           & 0.309                           & 0.275                           & 0.316    & 0.295                           & 0.321                           & {0.356} \\
\multicolumn{1}{c|}{}                                                   & NMI        & 0.331 & 0.159                           & 0.336                           & 0.221                           & 0.315    & 0.245                           & 0.316                           & 0.355                           \\
\multicolumn{1}{c|}{}                                                   & F1-score & {0.265}                           & 0.173                           & 0.246                           & 0.153                           & 0.251    & 0.193                           & 0.256                           & {0.265}                           \\ \hline
\multicolumn{1}{c|}{\multirow{3}{*}{Context}}                           & Purity     & 0.569                           & 0.425                           & 0.518                           & 0.354                           & 0.572    & 0.55                            & 0.61                            & {0.792} \\
\multicolumn{1}{c|}{}                                                   & NMI        & 0.557                           & 0.466                           & 0.407                           & 0.071                           & 0.515    & 0.300                             & 0.593                           & {0.655} \\
\multicolumn{1}{c|}{}                                                   & F1-score & 0.551                           & 0.487                           & 0.444                           & 0.235                           & 0.523    & 0.211                           & 0.573                           & {0.699} \\ \hline
\multicolumn{1}{c|}{\multirow{3}{*}{Epitope}}                           & Purity     & 0.627                           & 0.559                           & 0.559                           & {0.683} & 0.594    & 0.670                            & 0.656                           & 0.559                           \\
\multicolumn{1}{c|}{}                                                   & NMI        & 0.114                           & 0.044                           & 0.096                           & 0.093                           & 0.054    & 0.060                            & 0.126                           & {0.142} \\
\multicolumn{1}{c|}{}                                                   & F1-score & 0.587                           & {0.671} & 0.635                           & 0.585                           & 0.550     & 0.277                           & 0.582                           & 0.620                            \\ \hline
\multicolumn{1}{c|}{\multirow{3}{*}{Gene}}                              & Purity     & {1}     & 0.511                           & 1                               & 0.519                           & 0.935    & 0.965                           & 0.999                           & {1}     \\
\multicolumn{1}{c|}{}                                                   & NMI        & {1}     & 0.060                            & 0.994                           & 0.012                           & 0.782    & 0.158                           & 0.989                           & {1}     \\
\multicolumn{1}{c|}{}                                                   & F1-score & {1}     & 0.665                           & 0.999                           & 0.500                             & 0.914    & 0.059                           & 0.998                           & {1}     \\ \hline
\multicolumn{1}{c|}{\multirow{3}{*}{News}}                              & Purity     & 0.254                           & 0.210                            & 0.269                           & 0.241                           & 0.284    & {0.535} & 0.462                           & N.A                             \\
\multicolumn{1}{c|}{}                                                   & NMI        & 0.020                           & 0.002                           & 0.106                           & 0.007                           & 0.036    & {0.249} & 0.226                           & N.A                             \\
\multicolumn{1}{c|}{}                                                   & F1-score & 0.262                           & 0.253                           & 0.329                           & 0.218                           & 0.263    & 0.256                           & {0.344} & N.A                             \\ \hline
\multicolumn{1}{c|}{\multirow{3}{*}{Pioneer}}                           & Purity     & 0.798                           & 0.656                           & 0.652                           & 0.644                           & 0.662    & 0.713                           & {0.807} & 0.806                           \\
\multicolumn{1}{c|}{}                                                   & NMI        & {0.549} & 0.064                           & 0.175                           & 0.090                            & 0.097    & 0.181                           & 0.459                           & 0.474                           \\
\multicolumn{1}{c|}{}                                                   & F1-score & {0.688} & 0.657                           & 0.476                           & 0.406                           & 0.515    & 0.338                           & 0.652                           & 0.684                           \\ \hline
\multicolumn{1}{c|}{\multirow{3}{*}{Question}}                          & Purity     & 0.656                           & 0.518                           & 0.518                           & {0.745} & 0.604    & 0.570                            & 0.560                            & N.A                             \\
\multicolumn{1}{c|}{}                                                   & NMI        & 0.122                           & 0.022                           & 0.019                           & {0.295} & 0.081    & 0.047                           & 0.015                           & N.A                             \\
\multicolumn{1}{c|}{}                                                   & F1-score & 0.572                           & {0.666} & 0.619                           & 0.665                           & 0.591    & 0.546                           & 0.514                           & N.A                             \\ \hline
\multicolumn{1}{c|}{\multirow{3}{*}{Reuters}}                           & Purity     & 0.579                           & 0.255                           & 0.321                           & 0.347                           & 0.448    & 0.287                           & {0.699} & N.A                             \\
\multicolumn{1}{c|}{}                                                   & NMI        & 0.392                           & 0.097                           & 0.101                           & 0.048                           & 0.154    & 0.062                           & {0.482} & N.A                             \\
\multicolumn{1}{c|}{}                                                   & F1-score & 0.501                           & 0.298                           & 0.350                            & 0.292                           & 0.388    & 0.391                           & {0.582} & N.A                             \\ \hline
\multicolumn{1}{c|}{\multirow{3}{*}{Robot}}                             & Purity     & 0.639                            & 0.515                           & 0.544                           & 0.637                           & 0.557    & 0.535                           & 0.618                           & {0.683} \\
\multicolumn{1}{c|}{}                                                   & NMI        & 0.071                           & 0.046                           & 0.035                           & 0.056                           & 0.017    & 0.019                           & 0.068                           & {0.107} \\
\multicolumn{1}{c|}{}                                                   & F1-score & 0.551                          & {0.655} & 0.592                           & 0.538                           & 0.521    & 0.522                           & 0.564                           & 0.585                           \\ \hline
\multicolumn{1}{c|}{\multirow{3}{*}{Skating}}                           & Purity     & 0.208                           & 0.183                           & 0.222                           & 0.221                           & 0.220     & {0.232} & 0.228                           & 0.227                           \\
\multicolumn{1}{c|}{}                                                   & NMI        & 0.041                           & 0.029                           & 0.041                           & 0.023                           & 0.041    & {0.112} & 0.032                           & 0.053                           \\
\multicolumn{1}{c|}{}                                                   & F1-score & 0.224                            & {0.252} & 0.193                           & 0.150                            & 0.181    & 0.246                           & 0.164                           & 0.159                           \\ \hline
\multicolumn{1}{c|}{\multirow{3}{*}{Unix}}                              & Purity     & 0.506                           & 0.444                           & 0.491                           & 0.464                           & 0.479    & {0.756} & 0.451                           & N.A                             \\
\multicolumn{1}{c|}{}                                                   & NMI        & 0.110                             & 0.004                           & 0.094                           & 0.031                           & 0.055    & {0.224} & 0.018                           & N.A                             \\
\multicolumn{1}{c|}{}                                                   & F1-score & 0.385                           & {0.484} & 0.422                           & 0.308                           & 0.371    & 0.227                           & 0.438                           & N.A                             \\ \hline
\multicolumn{1}{c|}{\multirow{3}{*}{Webkb}}                             & Purity     & 0.502                           & 0.444                           & 0.443                           & 0.448                           & 0.479    & {0.723} & 0.473                           & N.A                             \\
\multicolumn{1}{c|}{}                                                   & NMI        & 0.094                           & 0.020                            & 0.055                           & 0.019                           & 0.072    & {0.199} & 0.126                           & N.A                             \\
\multicolumn{1}{c|}{}                                                   & F1-score & 0.452                           & 0.421                           & {0.455} & 0.378                           & 0.426    & 0.341                           & 0.403                           & N.A                             \\ \hline
\multicolumn{1}{c|}{\multirow{3}{*}{Average (Rank)}}   & Purity     & \textbf{0.584} (\textbf{2.5})                           & {0.421} (6.43)                            & {0.499} (4.43)                            & 0.474 (4.64)                            & 0.52 (3.71)     & 0.567 (3.21)                            & 0.577 (2.71)                            & /                               \\
 \multicolumn{1}{c|}{}                                                                        & NMI        & \textbf{0.33} (\textbf{2.14})                            & 0.077 (6.07)                            & 0.2 (3.43)                            & 0.081 (5.5)                            & 0.185 (4)     & 0.151 (3.79)                            & 0.274 (2.93)                            & /                               \\
\multicolumn{1}{c|}{}                                                                        & F1-score & \textbf{0.525} (\textbf{2.5})                            & 0.478 (2.93)                            & 0.476 (3.21)                            & 0.367 (5.79)                            &0.455 (4.14)     &0.303 (5.93)                            & 0.496 (3.5)                            & /                              \\\hline

\end{tabular}
\label{performance_comparison}
\end{table*}

\begin{figure}[!t]%
    \centering
    \includegraphics[scale = 0.35]{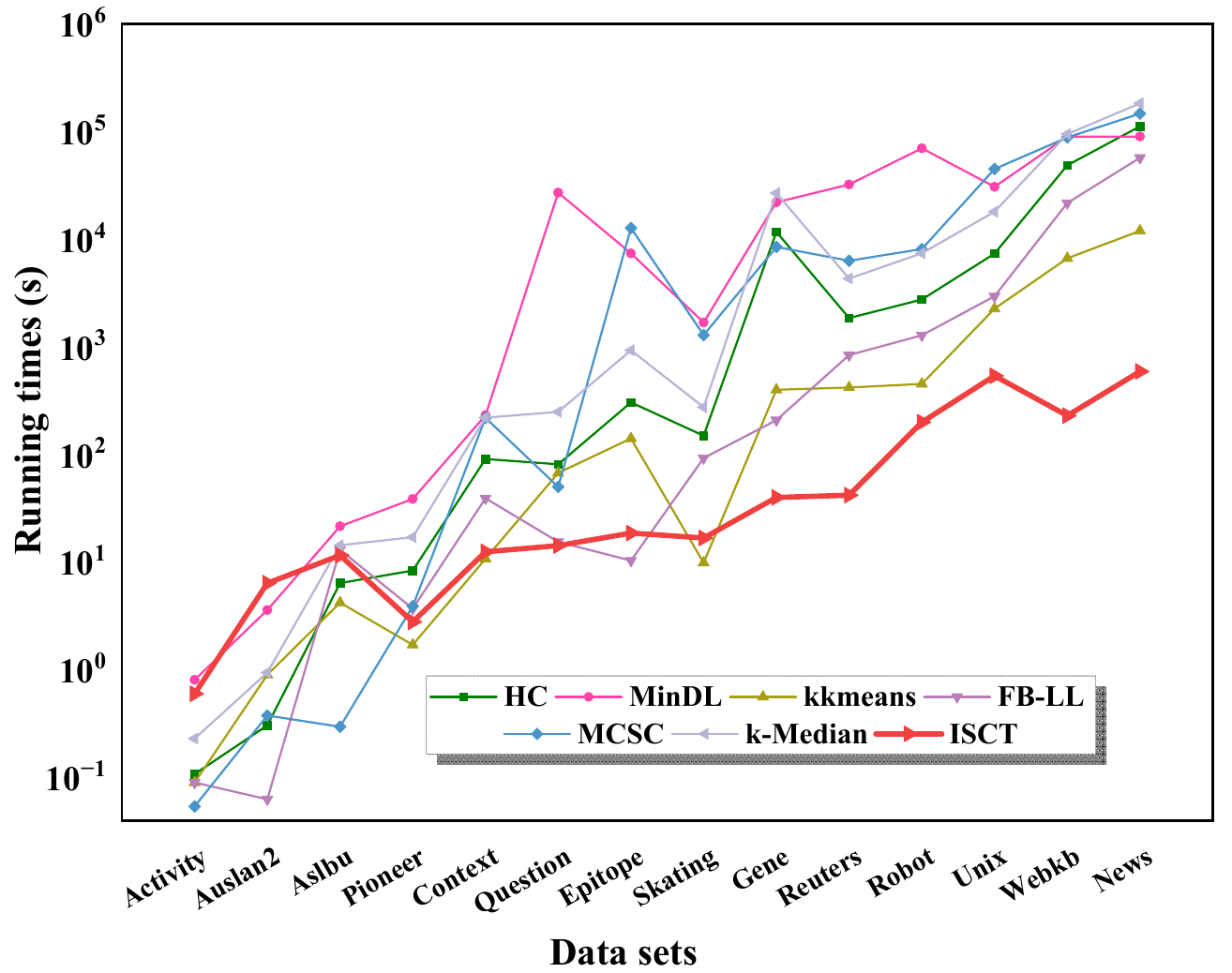}
    \caption{Running time comparison of ISCT, HC, FB-LL, MCSC, $k$-Median, MinDL, kkmeans, and SGT.}
    \label{time}
\end{figure}

Our method offers a highly interpretable structure for the sequence clustering process. However, it is important to compare our algorithm with existing algorithms as well. The performance comparison results in terms of Purity, NMI, and F1-score are presented in Table \ref{performance_comparison}. Due to memory limitations during the experiment, some results of SGT were not available. Figure \ref{time} shows the running time of the different methods on the 14 data sets. 

\begin{figure*}[!htbp]
    \centering
    \begin{subfigure}{0.3\textwidth}
        \includegraphics[scale=0.2]{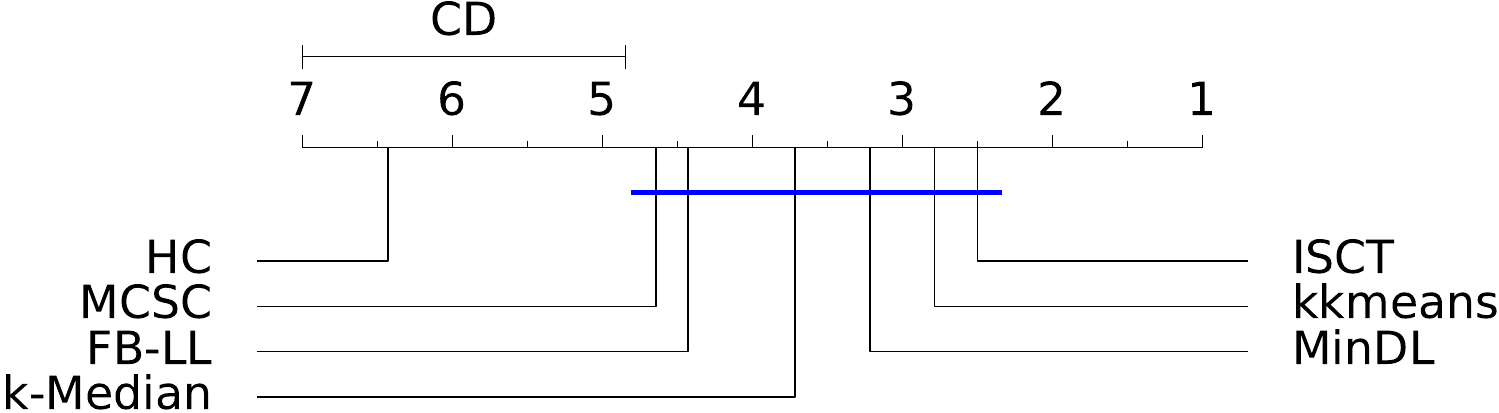 }
        \caption{Purity}

    \end{subfigure}
    \begin{subfigure}{0.3\textwidth}
        \includegraphics[scale=0.2]{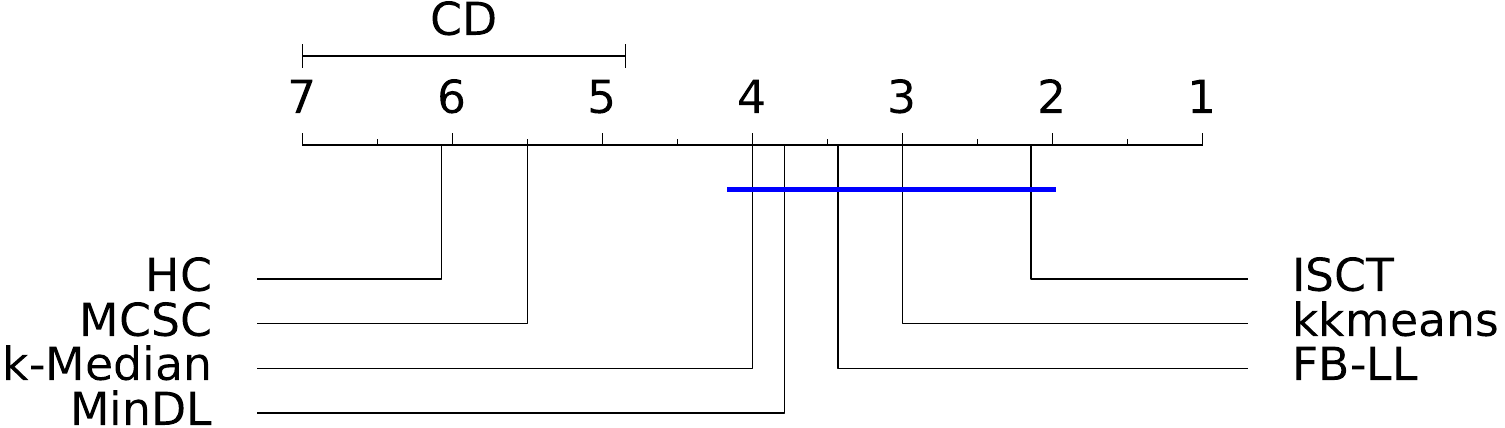}
        \caption{NMI}
    \end{subfigure}
    \begin{subfigure}{0.3\textwidth}
        \includegraphics[scale=0.2]{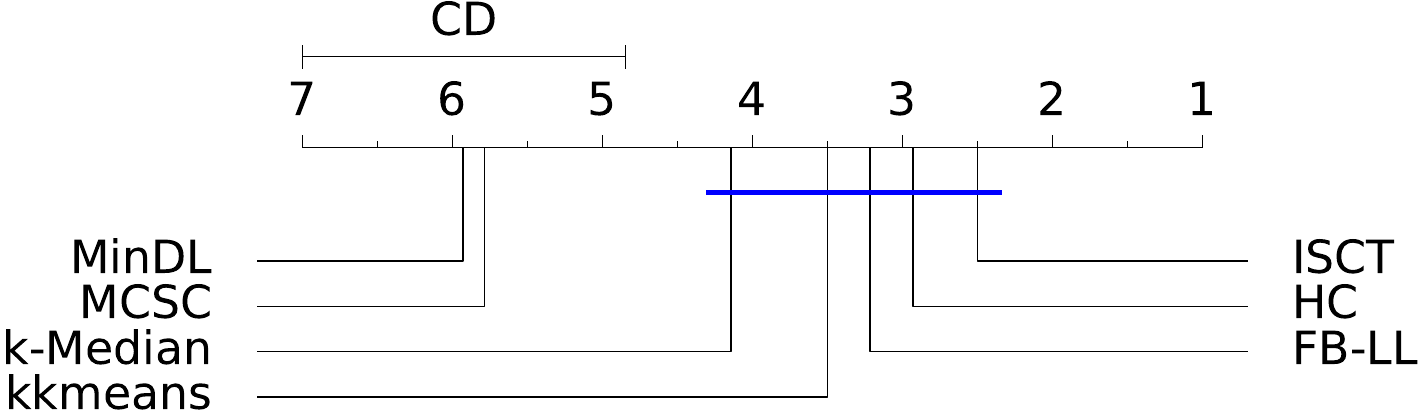}
        \caption{F1-score}
    \end{subfigure}
    \caption{Bonferonni-Dunn critical difference diagrams at a significance level of 0.05.}
    \label{Bonferonni-Dunn}
\end{figure*}

Our proposed ISCT method achieves the best average performance across all the evaluation measures, indicating that it outperforms the other methods in terms of identification performance. To further evaluate ISCT's performance compared to other methods, Bonferroni-Dunn tests were conducted to validate the null hypothesis that ISCT performs similarly to the other methods. The significance testing results are presented in Figure \ref{Bonferonni-Dunn}. If there are no connections between our method and the existing algorithm, it indicates that our method is significantly better. From the results, it can be observed that while ISCT does not significantly outperform all other methods, it does exhibit better performance than HC. It is important to note that MinDL achieves the highest purity on certain data sets. However, it often generates a large number of clusters with only a few sequences in each cluster. For instance, on the ``Pioneer" data set, MinDL produces $23$ clusters, whereas the ground truth number of clusters is only $3$. As for SGT, it performs well on many datasets but struggles with data sets containing large item sets due to its high storage space requirements. In comparison to two hierarchical clustering algorithms (HC and MCSC), our method consistently outperforms them on the majority of data sets.

When comparing ISCT with the feature-based method FB-LL, ISCT outperforms FB-LL in terms of purity and NMI on all data sets, and F1-score on 9 data sets. This improvement is mainly attributed to the pattern mining strategy employed by ISCT. In FB-LL, the feature vector is generated by mining frequent patterns. However, frequent patterns are often observed throughout the entire data set, which means that the transformed feature vectors for different sequences might be similar. As a result, the mined patterns rarely provide distinctive information for clustering. On the other hand, by employing random projection and a similarity measure based on the LCS, the transformed features become more informative for clustering. This approach enhances the discriminative power of the feature vectors, enabling better differentiation between different sequences during the clustering process. 

When comparing ISCT with kkmeans and $k$-Median, ISCT demonstrates superior performance on the majority of data sets. Additionally, as depicted in Figure \ref{time}, our method exhibits significantly improved efficiency. In summary, ISCT achieves comparable performance to state-of-the-art methods while delivering considerably faster computation times and ensuring interpretability.



\subsection{Investigation of RQ2}
The cluster membership can be viewed as pseudo labels. A naive approach to utilize cluster membership as the class label to train a classifier using popular tree-based methods. However, the interpretability of these tree-based methods is very good, as it is difficult to control the size of the resulting tree. In practice, it is often observed that a large tree is generated, which poses challenges in understanding the decision process.

In this section, we compare the interpretability of our ISCT method with SeqDT \cite{9411677}, as both methods use patterns as the splitting points. We aim to assess which method provides a more interpretable representation of the clustering process, of which the interpretability is measured with the constructed tree size. For SeqDT, we use different parameters on the threshold of Gini impurity to show its influence on the size of the tree.

\begin{table}[!t]
\centering
\caption{Comparison of average tree sizes between ISCT and SeqDT in terms of the number of leaf nodes with different Gini impurities (averaged over ten experiments).}
\label{treesize}
\begin{tabular}{|c||c|c|c|c|}
\hline
Datasets  & Ground Truth  & \textbf{ISCT} & SeqDT\_0.1 & SeqDT\_0.01 \\ \hline\hline
Activity & 2       & \textbf{2}    & \textbf{2}             & \textbf{2}              \\ \hline
Aslbu    & 7       & \textbf{7}    & 32.2          & 33.7           \\ \hline
Auslan2  & 10      & \textbf{9.5}  & 11.9          & 13.1           \\ \hline
Context  & 5       & \textbf{5}    & 9.5           & 10.1           \\ \hline
Epitope  & 2       & \textbf{2}    & \textbf{2}             & 11.6           \\ \hline
Gene     & 2       & \textbf{2}    & \textbf{2}             & \textbf{2}              \\ \hline
News     & 5       & \textbf{5}    & 94.1          & 159.3          \\ \hline
Pioneer  & 3       & \textbf{3}    & 4.6           & 3.7            \\ \hline
Question & 2       & \textbf{2}    & 6.8           & 4.1            \\ \hline
Reuters  & 4       & \textbf{4}    & 34.5          & 35.7           \\ \hline
Robot    & 2       & \textbf{2}    & 76.1          & 107.2          \\ \hline
Skating  & 7       & \textbf{7}    & 71.4          & 66.2           \\ \hline
Unix     & 4       & \textbf{4}    & 53.3          & 84.5           \\ \hline
Webkb    & 3       & \textbf{3}    & 132.1         & 185.6          \\ \hline
\end{tabular}
\end{table}

The results of the tree size comparison in terms of the number of leaf nodes between ISCT and SeqDT with different Gini impurities are shown in Table \ref{treesize}. The ground truth column represents the actual number of clusters $|C|$ in each data set. 

For most data sets, SeqDT with a threshold of 0.01 produces larger trees compared to SeqDT with a threshold of 0.1. This suggests that a lower threshold leads to more complex trees, which may be harder to interpret. However, even with a threshold of 0.1 and employing the PEP (Pessimistic Error Pruning) strategy, SeqDT still generates larger trees compared to ISCT, indicating its limited interpretability.

In contrast, ISCT consistently reports the same number of leaf nodes as the ground truth, except on the ``Auslan2" data set where there are a significant number of duplicate sequences marked as different classes. Excluding this exceptional case, the consistent number of leaf nodes in ISCT indicates the effectiveness of our tree construction strategy. It demonstrates that ISCT can accurately capture the underlying clustering structure while maintaining a concise and intuitive decision tree structure. As a result, ISCT facilitates a better understanding and provides valuable insights into the clustering results.

\subsection{Investigation of RQ3}


During the tree construction process, we introduce a boosting approach to facilitate the construction. In order to validate the efficacy of our proposed strategy, we perform experiments on datasets that consist of more than two categories.

\begin{table}[!t]
\centering
\caption{Performance comparison on different tree construction strategies in terms of average reduction rate with respect to Purity, NMI and F1-score.}
\label{averageboostonicst}
\begin{tabular}{|c||c|c|} 
\hline

Method &  \textbf{ICST(Boost)} & ICST(Direct)                    \\ 
\hline\hline
Purity &  \textbf{-0.099}      & -0.158                      \\ 
\hline
NMI    &  \textbf{-0.202}       & -0.264                     \\ 
\hline
F1     &  \textbf{0.040}          & 0.039                    \\
\hline
\end{tabular}
\end{table}



Table \ref{averageboostonicst} presents a comparison of the average performance reduction rates between different tree construction methods and the random projection clustering algorithm. The results demonstrate that the performance reduction rates of boosting strategy are superior to direct construction method.

\begin{figure}[!t]%
    \centering
    \includegraphics[scale = 0.22]{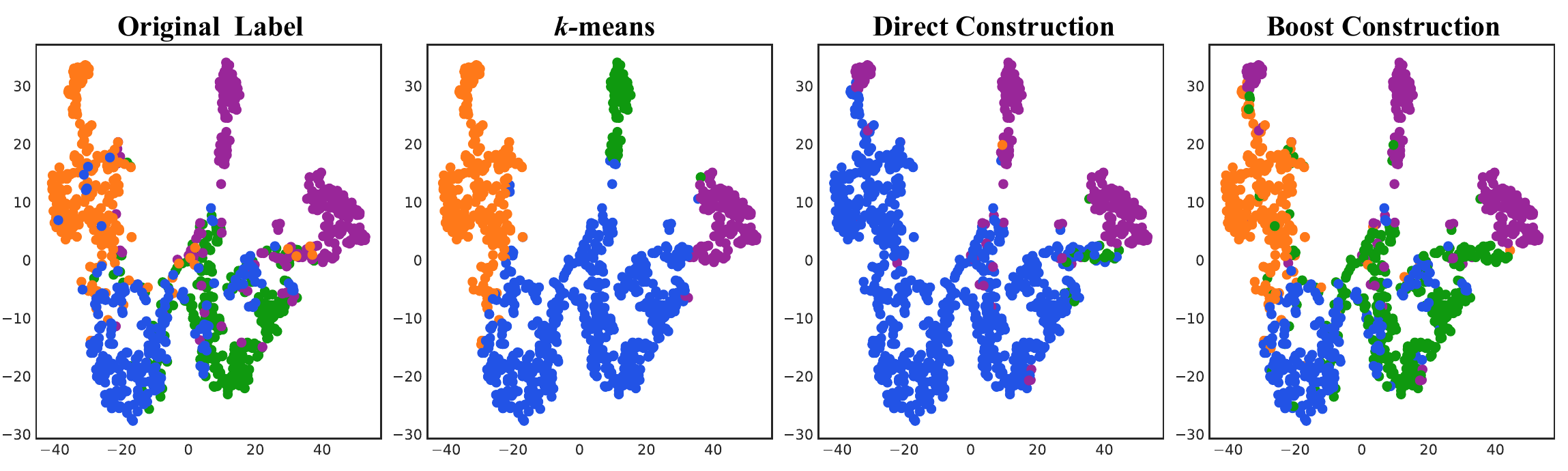}
    \caption{T-SNE visualization of feature space with original label, $k$-means, direct and boost constructions. }
    \label{tsne}
\end{figure}

To show the benefits of our construction strategy, we employ t-SNE to visualize the ``Reuters" as an illustrative example. The visualization results are presented in Figure \ref{tsne}. It is evident from the figure that the distinguishability of the four clusters in the random projection subspace is not particularly apparent. Consequently, this leads the $k$-means algorithm to erroneously merge the blue and green clusters into a single cluster. Constructing a tree based on this partition would result in the identification of inappropriate patterns. However, by leveraging our boosting strategy, we successfully extract patterns that accurately differentiate between the four distinct categories.

\section{Conclusion}
\label{conclusions}
This paper introduces the Interpretable Sequence Clustering Tree (ISCT), which aims to construct a concise and interpretable decision tree for sequence clustering. The ISCT consists of $k$ leaf nodes that correspond to $k$ clusters, enabling a transparent and comprehensible decision-making process for sequence clustering. Experimental results on 14 real-world benchmark data sets demonstrate that our proposed ISCT method achieves state-of-the-art performance while ensuring interpretability.


\bibliographystyle{IEEEtran}
\bibliography{myrefs}{}

\end{document}